\documentclass[letterpaper, 10 pt, conference]{ieeeconf}  

\IEEEoverridecommandlockouts                              %
\usepackage{graphics} 
\usepackage{epsfig} 
\usepackage{mathptmx} 
\usepackage{times} 
\usepackage{amsmath} 
\usepackage{amssymb}  
\usepackage{booktabs}
\usepackage{etoolbox}
\makeatletter
\patchcmd{\@makecaption}
  {\scshape}
  {}
  {}
  {}
\makeatletter
\patchcmd{\@makecaption}
  {\\}
  {.\ }
  {}
  {}
\makeatother

\title{\LARGE \bf
Dyna-DepthFormer: Multi-frame Transformer \\ for Self-Supervised Depth Estimation in Dynamic Scenes
}

\author{Songchun Zhang and Chunhui Zhao
\thanks{*This work is supported by The National Science Fund for Distinguished Young Scholars (No. 62125306), and the Research Project of the State 
Key Laboratory of Industrial Control Technology, Zhejiang University, China (ICT2021A15). (The corresponding author is Chunhui Zhao)}
\thanks{Songchun Zhang and Chunhui Zhao are with the State Key Laboratory of Industrial Control Technology, College of Control Science and Engineering, Zhejiang University, Hangzhou 310027, China}%
}

\begin{document}

\maketitle
\thispagestyle{empty}
\pagestyle{empty}

\begin{abstract}

Self-supervised methods have showed promising results on depth estimation task. 
However, previous methods estimate the target depth map and camera ego-motion simultaneously, underusing multi-frame correlation information and ignoring the motion of dynamic objects.
In this paper, we propose a novel Dyna-Depthformer framework, which predicts scene depth and 3D motion field jointly and aggregates multi-frame information with transformer.
Our contributions are two-fold.
First, we leverage multi-view correlation through a series of self- and cross-attention layers in order to obtain enhanced depth feature representation.
Specifically, we use the perspective transformation to acquire the initial reference point, and use deformable attention to reduce the computational cost.
Second, we propose a warping-based Motion Network to estimate the motion field of dynamic objects without using semantic prior.
To improve the motion field predictions, we propose an iterative optimization strategy, together with a sparsity-regularized loss.
The entire pipeline achieves end-to-end self-supervised training by constructing a minimum reprojection loss.
Extensive experiments on the KITTI and Cityscapes benchmarks demonstrate the effectiveness of our method and show that our method outperforms state-of-the-art algorithms.

\end{abstract}

\section{INTRODUCTION}

Depth estimation is a fundamental computer vision task, which plays an influential role in many practical applications, such as  autonomous driving, robot grasping and augmented reality.
The traditional geometric approaches~\cite{schonberger2016colmap, gherardi2010improvingsfm, mur2015orb} solve this problem by exploiting the key point matching relationship between adjacent frames and recovering the depth information of 2D pixels through triangulation.
Although traditional geometric methods have achieved promising performance, they are unable to handle low-texture regions and non-stationary scenes.
In addition, the huge computational cost of feature matching limits their practical application.

Recently, deep learning techniques have improved this research area by training networks to directly predict depth from single image.
Despite the fact that supervised learning methods have achieved good results~\cite{adabins, li2022depthformer, simipu}, the cost of labeling ground truth limits the practical application of such methods.
To solve this problem, unsupervised depth estimation~\cite{monodepth2,manydepth,epc++,zhou2017unsupervised,zhan2018unsupervised} is gradually becoming a novel research trend, which treats depth estimation as a view synthesis problem.
Although self-supervised depth estimation has made great progress in recent years, there are still some unresolved issues. 
First, without the feature matching step, the unsupervised depth estimation cannot leverage the semantic information from adjacent frames, which leads to inaccurate depth estimation results.
Second, it is known that the photometric loss is sensitive to dynamic objects, which violates the assumption of the static world. Therefore, the perspective projection will be non-rigid, resulting in unstable training process.
On the one hand, Manydepth\cite{manydepth} leveraged cost-volume based method to aggregate multi-frame information. 
However, inaccurate pose estimation introduces additional noise to feature matching, which can sharply decreasing model performance in dynamic scenes.
On the other hand, recent work have tried to introduce instance segmentaion \cite{lee2019instance}, auto-mask \cite{monodepth2} and optical flow \cite{yin2018geonet} to deal with dynamic objects under the self-supervised learning framework. 
These methods introduce expensive additional labeling costs, which are therefore difficult to be applied in real-world applications.

\begin{figure}[t!]
    \centering
    \includegraphics[width=0.35\textwidth]{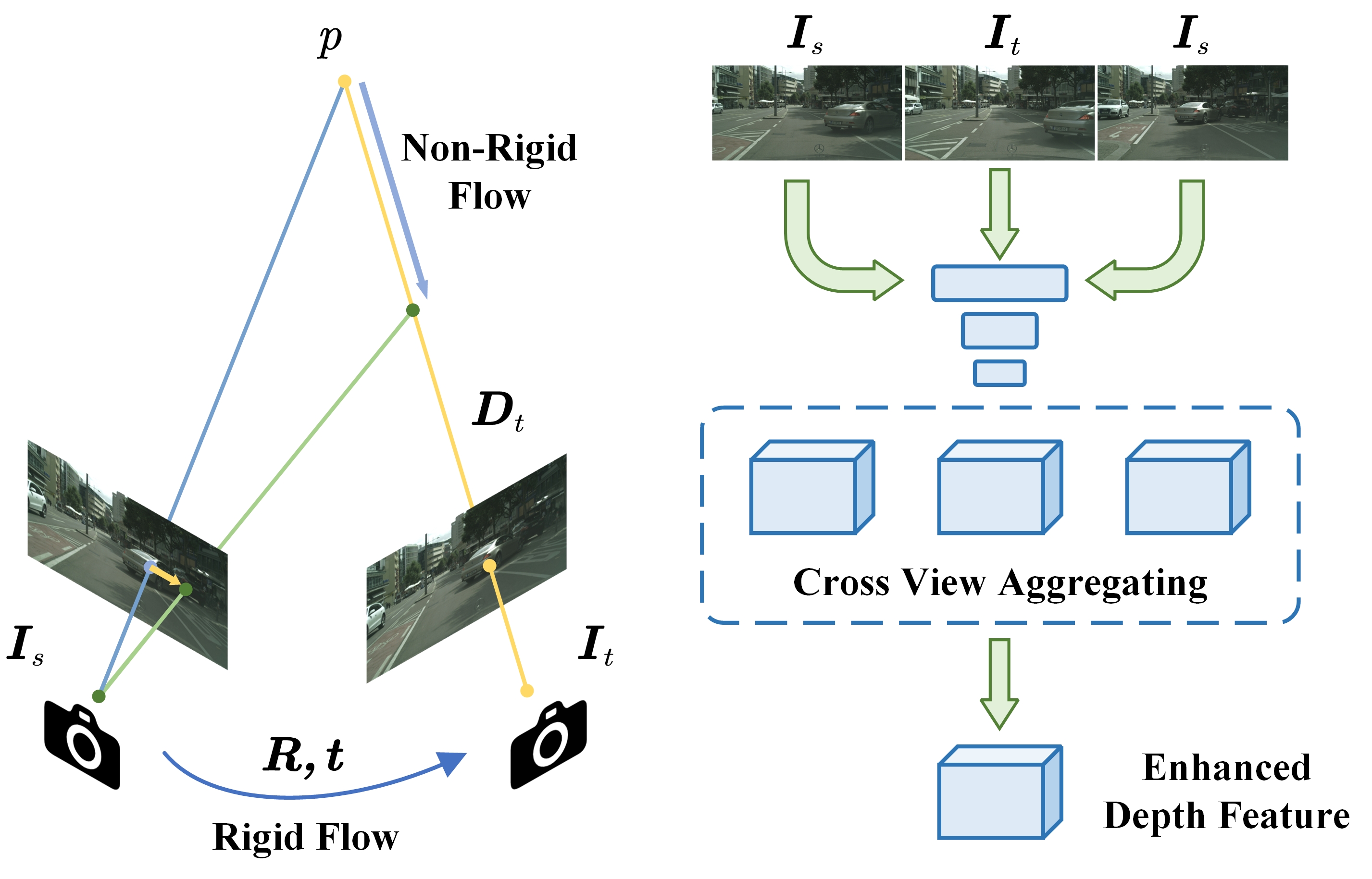}
    \caption{
        \textbf{Overall architecture} of our proposed method. On the one hand, we model the dynamic objects in the scene to improve the accuracy of depth estimation in dynamic regions. On the other hand, considering the presence of dynamic objects, we design a geometry-based cross-view feature fusion approach to obtain discriminative feature representations.
    }
    \label{fig:overview}
\end{figure}

In this work, we propose a \textbf{Dyna-DepthFormer} network in order to solve the above-mentioned problems. 
Specifically, we propose a depth feature enhance module consisting of self-attention and cross-attention module to capture the correlation in multi-view feature map.
Inspired by \cite{zhu2020deformable}, we realize this module by constructing a Transformer-based module, while the deformable attention is adopted to reduce the computational costs and adjust sample point to capture the feature in the local region.
The generated features will be more discriminative and robust, since it focus on the surrounding local context and aggregates relevant information.
Furthermore, we propose a novel Motion Network, which model the dynamic object by leveraging the motion field prediction. Then, We can use the motion field and camera motion to calculate the reprojection loss without using semantic prior. In summary, our contributions are:




\begin{itemize}
  \item 
  We propose a transformer-based Depth Network, which enhances the depth feature representation through multi-view feature aggregating. Considering the high computation of dense attention, we propose a perspective-projection based temporal deformable cross-attention module that efficiently reduces the computation.
  \item
  In addition to camera ego motion, we proposed to estimate the 3D motion field of moving objects, and improve its completeness through iterative refinement. This eliminates the necessity of the scene-static assumption and enables unsupervised method to use more unlabeled data for training.
  \item
  We demonstrate that our proposed method achieves promising results in monocular depth estimation on benchmark datasets like KITTI and Cityscapes while reducing computing cost.
\end{itemize}

\section{RELATED WORK}

\subsection{Self-Supervised Depth Estimation}
Self-supervised depth estimation methods, in contrast to supervised depth estimation methods, can estimate depth from unlabeled video training data.
The work of Monodepth~\cite{godard2017unsupervised} introduced a self-supervised depth estimation method by leveraging the left-right consistency. 
SFM-learner~\cite{zhou2017unsupervised} estimated the relative poses between adjacent frames simultaneously with the addition of a pose network.
Then, Monodepth2~\cite{monodepth2} absorbed the advantages of the prior methods and proposed a universal framework, which can be trained with stereo pairs and videos data.
However, this approach was afflicted by issues such as scale ambiguity, dynamic objects, low-texture areas, and non-Lambertian surfaces.
In recent years, several strategies have been developed to improve the performance of depth estimation by tackling the aforementioned difficulties.
Featdepth~\cite{shu2020featdepth} introduced an auto-encoder structure in order to obtain a robust feature representation, which can improve the depth estimation accuracy in the low-texture regions.
Packnet~\cite{packnet} proposed a novel 3D convolution packing module to overcome the scale ambiguity problem, while it suffered from heavy computational cost.

With the development of cost-volume based multi-view stereo (MVS) methods, some of the work~\cite{manydepth, guizilini2022multi, wang2021deep, monorec} introduced the cost-volume to self-supervised depth estimation field, which enabled the learning of additional geometric cues from feature matching across frames.
However, due to the hand-crafted similarity metrics, the cost volume will be an unreliable source of depth information in locations where objects are moving or surfaces are untextured.

\subsection{Attention Mechanisms in Depth Estimation}
Self-Attention mechanism have shown promising results in the field of natural language processing~\cite{attention_all} and are becoming increasingly popular in computer vision region ~\cite{dosovitskiy2020vit, liu2021swin}. 
In the field of depth estimation, current methods can be divided into two categories. 
One class of methods apply attention mechanisms to exploit long-range correlation and local information of single view feature representation representation~\cite{lyu2021hrdepth,yan2021channel,li2022depthformer}.
The other class of methods tried to fuse features from multiple sources through an attention mechanism.
Some work \cite{jung2021fsre, li2021learning} leveraged semantic information to guide depth feature representation.
More related to our work, \cite{ruhkamp2021attention} proposed a spatial-temporal attention module to fuse cross-view information through two cascaded attention blocks.
However, their approach did not make sufficient use of geometric information, resulting in high computational cost.

\subsection{Depth estimation in Dynamic Environment}
Video sequence data is usually captured in scenes containing dynamic objects. 
However, the non-rigid parts of the scene can affect the self-supervised training. 
To alleviate this problem, \cite{monodepth2} introduced an novel auto-mask method to exclude dynamic object areas. 
Although auto-mask can exclude dynamic object pixels, it also masks away some pixels that are useful for training.
Some of the work \cite{boulahbal2022instance, lee2019instance} leveraged semantic information to seperate the foreground and background object.
Then, they estimated the object motion and camera ego-motion seperately.
However, their methods require expensive labeling costs.
Recent work attempted to model the moving object explicitly with optical flow~\cite{yin2018geonet} or 3D motion field \cite{li2020dynamic, lee2021attentive}. 
Comparing to \cite{lee2021attentive}, we design a one-stage warping based network to estimate the motion field, which directly represent object motion in 3D space.
Furthermore, to improve the quality of the predicted motion fields, we also propose an iterative optimization methods with sparsity regularization loss.

\section{METHOD}

In this section, we propose Dyna-DepthFormer, a self-supervised depth estimation Framework with Transformer.
We first  describe the problem formulation, followed by a framework overview of our method in Section \ref{sec:overview}.
In order to predict accurate depth map of the target image, we present a depth estimation network in Section \ref{sec:depth_estimate}, which can fuse multi-view geometric information efficiently.
In Section \ref{sec:motion_estimation}, we give details on the Motion Estimation Network, an module designed to deal with the moving objects and non-rigid region in training process. 
In Section \ref{sec:loss_function}, we further elaborate on the self-supervised loss function and regularization to finalize our training strategy.

\subsection{Framework Overview}
\label{sec:overview}
\begin{figure*}[t!]
\centering
\includegraphics[width=0.60\textwidth]{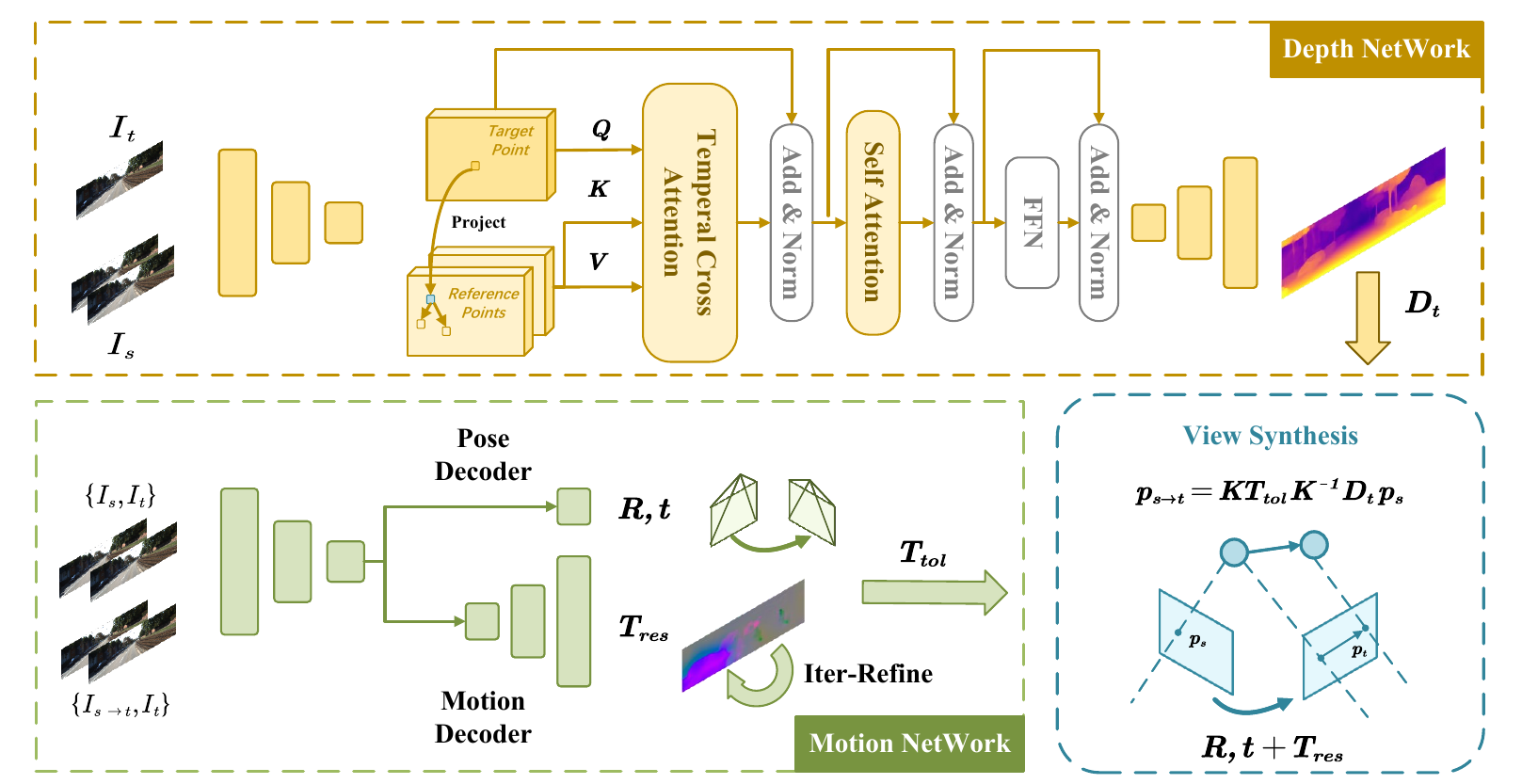}
\caption{
\textbf{Pipeline of our proposed Dyna-Depthformer architecture}. The entire pipeline includes three parts, Depth Network, Motion Network and View Synthesis Module.  For Depth Network, cross-view attention and self attention module are adopted to enhance the feature modeling and depth inferring. For Motion Network, we predict ego motion and object motion with two decoupled head. The view synthesis module reconstructs the image by differentiable warping as a way to perform self-supervised learning.
}
\label{fig:model_pipeline}
\vspace{-4mm}
\end{figure*}

The overall framework, as illustrated in Fig \ref{fig:model_pipeline}, consists of four parts: the pre-trained backbone, the depth estimation network, the motion estimation network and the view synthesis module. 

The pre-trained backbone takes the scaled images pair $\{\mathbf{I}_t,\mathbf{I}_{s}\}$ as input to generate multi-scale image feature. We first use motion network $\mathbf{M}_{\psi}$ and pose network $\mathbf{P}_{\phi}$ to generate ego motion $\{\mathbf{R}_{t \rightarrow s},\mathbf{t}_{t \rightarrow s}\}$ and object motion field $\mathbf{T}_{res}$. Then, we perform perspective projection based attention to obtain enhanced multi-view depth feature to generate accurate depth prediction $\mathbf{D}_t$ with depth network $\mathbf{D}_{\theta}$. 

\begin{equation}
    \begin{gathered}
\left\{\mathbf{R}_{t \rightarrow s}, \mathbf{t}_{t \rightarrow s}\right\}=\mathbf{P}_\phi\left(\left\{\mathbf{I}_t, \mathbf{I}_s\right\}\right) \\
\mathbf{T}_{\text {res }}=\mathbf{M}_{\psi}\left(\left\{\mathbf{I}_t, \mathbf{I}_{s \rightarrow t}\right\}\right) \\
\mathbf{D}_t=\mathbf{D}_\theta\left(\mathbf{I}_t\right).
\end{gathered}
\end{equation}

With the predicted depth $\mathbf{D}_t \in \mathbb{R}^{1 \times H \times W}$, relative camera pose $\{ \mathbf{R}_{t \rightarrow s}, \mathbf{t}_{t \rightarrow s} \}$ and object motion field $\mathbf{T}_{res}\in\mathbb{R}^{3 \times W \times H}$, we can synthesize the target frame $\mathbf{I}_t$ using the adjacent source frame $\mathbf{I}_s$ by differentiable view synthesis module~\cite{packnet}.
The difference between the source image and the reconstructed image can provide supervisory signal for self supervised training.

\subsection{Depth Estimation Network}
\label{sec:depth_estimate}
The goal of the depth estimation network ${D}_{\theta}$ is to map the input RGB images $\mathbf{I}_t$ to depth map $\mathbf{D}_t$. 
Noted that, monocular depth estimation is inherently an ill-posed problem, which suffers from moving object, textureless region and non-rigid region, etc.
Many previous work~\cite{yan2021channelwise, lyu2021hrdepth,shu2020featdepth} attempted to improve feature representation by applying self-attention module to current picture feature in order to address the aforementioned issue.
These approaches, however, have substantial computational costs and do not leverage geometric previous information.
Instead, we propose an perspective-project-based temporal cross-attention module to aggregate features in a learnable way.

\textbf{Feature Extract.} Given a target image $\mathbf{I_t} \in \mathbb{R}^{3 \times H \times W}$ and its neighboring source images $\mathbf{I_s} \in \{ \mathbf{I_{t+1}},\mathbf{I_{t-1}} \}$, the first step is to extract the multi-scale image features of these images. 
A shared backbone~\cite{he2016resnet} is applied, where the images are downscaled $N$ times to produce multi-scale deep features map $\mathbf{F}_{i=\{t-1, t, t+1\}}^{k=0,\ldots,N-1} \in \mathbb{R}^{C_k \times \frac{H}{2^k} \times \frac{W}{2^k}}$.
Then, we feed the extracted feature maps to the subsequent module.

\textbf{Temporal Cross-Attention.} 
As single-view depth estimation suffers from ill-posed problem, our method enhance the feature representation of target image by aggregating its neighboring image features with cross attention mechanisms.

\begin{figure}[t!]
    \centering
    \includegraphics[width=0.3\textwidth]{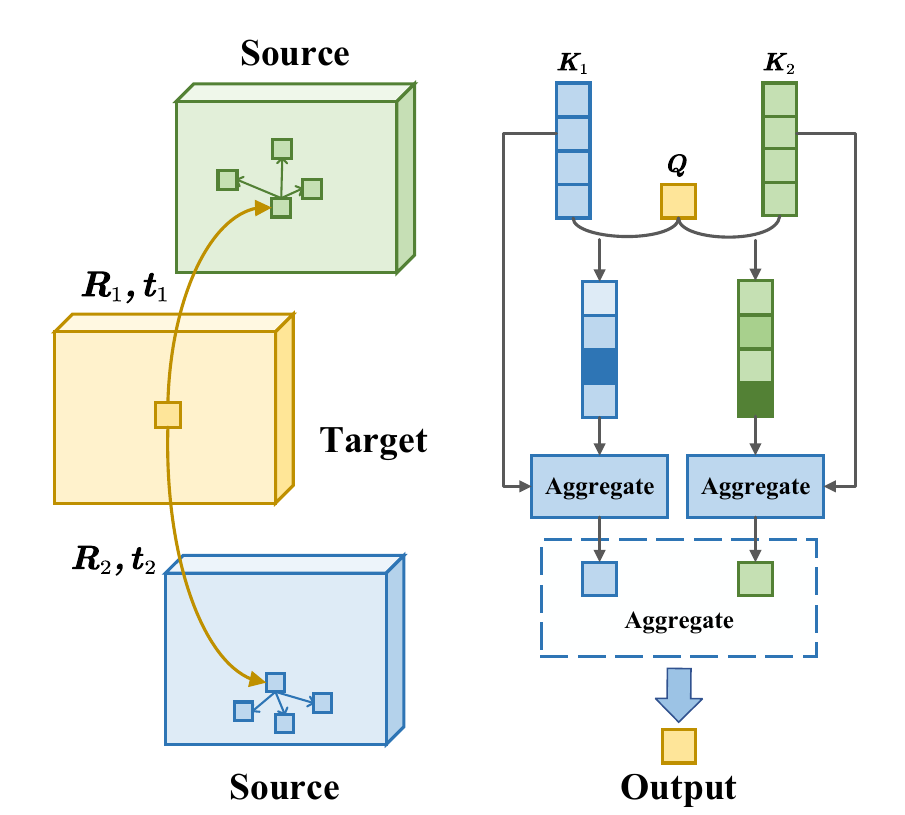}
    \caption{
        The details of a \textbf{Temporal Cross Attention Module.} This module fuses adjacent frame features by deformable cross-attention mechanism to obtain a robust and discriminative depth feature representation.
    }
    \label{fig:cross_attention}
    \vspace{-4mm}
\end{figure}

We use the features of the target image as \textit{Query} to match the features in the source image (\textit{Keys}). 
However, due to the huge number of key-value pairs, the original attention mechanism cannot be applied directly. 
Meanwhile, the epipolar sampling method~\cite{wang2022mvster,li2022depthformer} would suffer from camera parameter noise or scale-ambiguous problem as a result of the inaccurate pose estimate.
Inspired by \cite{zhu2020deformable, li2022bevformer}, we obtained the key-value pairs in the neighboring image features with a learnable way to reduce computational cost and increase efficiency.
Fig.~\ref{fig:cross_attention} illustrates the feature warping process and the generation of key-value pairs in our cross-attention module. 
Given a query point $\mathbf{p}_t=(u,v)$ in $\mathbf{F}_t$, it can be warped to the source camera coordinate by:

\begin{equation}
    \left(\mathbf{p}_s~1 \right)^T = \mathbf{K}\left(\mathbf{R}_{t \rightarrow s} \hat{\mathbf{D}}_t\left(\mathbf{p}_t\right) \mathbf{K}^{-1}\left (\mathbf{p}_t~1)\right)^T+\mathbf{t}_{t \rightarrow s}\right),
\end{equation}
where $\mathbf{p}_s$ is the associated point of the query point on the source plane, $\mathbf{K}$ is the camera intrinsics given in advance,  $\{\mathbf{R}_{t \rightarrow s},\mathbf{t}_{res}\}$ indicates the relative camera pose predicted by motion network, and $\hat{\mathbf{D}}_t(p_t)$ is the depth value at point $\mathbf{p}_t$. 
Note that, we use an auxiliary single scale depth decoder to generate a coarse predicted initial depth estimate $\hat{\mathbf{D}}_t$.

Then, we search for the most relevant features of the query feature in the neighborhood of $\mathbf{p}_s$.
Specifically, we learn some offsets $\Delta \mathbf{p} \in \mathbb{R}^2$ to the point $\mathbf{p}_s$ with MLP layer.
The deformable attention feature is calculated by:

\begin{equation}
    \mathbf{F}_t^{\prime}(\mathbf{p}_t)=\sum_{m=1}^M \mathbf{W}_m\left[\sum_{k=1}^K A_{m  k} \cdot \mathbf{W}_m^{\prime} \mathbf{F}_s\left(\mathbf{p}_s+\Delta \mathbf{p}_{m k}\right)\right],
\end{equation}
where $m$ indexes the attention head, $k$ indexes the sampled keys, $K$ is the total sampled points, $\mathbf{F}^{\prime}_t$ indicates the enhanced target feature, and $\mathbf{A}_{mk}$ indicates the scalar attention weight between \textit{Query} and \textit{Key} embedding. 
As $\mathbf{p}_s+\Delta \mathbf{p}_{m k}$ can be a fraction, bilinear interpolation is used to compute $\mathbf{F}_s\left(\mathbf{p}_s+\Delta \mathbf{p}_{m k}\right)$.
Note that $\mathbf{p}_s$ can also be regarded as an coarse initial point generated by the geometric prior, which will be optimized in the subsequent training process.
After that, We can fuse multiple aggregated features from source views by linear projection. 


\textbf{Self-Attention Refinement.} 
The enhanced feature $F_t^{\prime}$ are generated following temporal cross-view feature aggregation, which is more robust, but matching noise is maintained. 
Inspired by the cost aggregation step in the traditional stereo matching task~\cite{sgm}, we consider leveraging self-attention to refine the local features. 
The vectors in the feature map aggregate the discriminant features in the neighborhood based on cosine similarity to smooth the noisy information in cross-view enhanced feature map.
Specifically, Given the query point $\mathbf{p}_q=(u,v)$, the query feature $z_q$, and the enhanced feature $\mathbf{F}_t^{\prime}$, the self-attention refined feature is calculated by:
\begin{equation}
    \mathbf{F}_t^{\prime\prime}(\mathbf{p}_q)=\sum_{m=1}^M \mathbf{W}_m\left[\sum_{k=1}^K A_{m  k} \cdot \mathbf{W}_m^{\prime} \mathbf{F}_t^{\prime}\left(\mathbf{p}_q+\Delta \mathbf{p}_{m k}\right)\right].
\end{equation}

After obtaining the enhanced feature $\mathbf{F}_t^{\prime\prime}$, we utilize a straightforward decoder mentioned in~\cite{lyu2021hrdepth} to generate the multi-scale depth predictions $\mathbf{D}_t$.

\begin{table*}[t!]
\vspace{2mm}
\centering
\caption{\textbf{Depth estimation results} on the KITTI and Cityscapes datasets. \emph{Multi-Fr.} indicates using multiple frames during test time. \emph{Semantic.} indicates leveraging semantic prior during training process. K and CS denote the KITTI and the Cityscapes datasets.}
\label{tab:depth_result}
\scalebox{0.8}{
\begin{tabular}{lcccccccccc}
\toprule
\textbf{Method} & Multi-Fr & Semantic & Dataset & AbsRel & SqRel & RMSE & RMSE$_{log}$ & $\delta < 1.25$ & $\delta < 1.25^2$ & $\delta < 1.25^3$ \\ \midrule
Struct2depth~\cite{struct2depth} &  & \checkmark & K & 0.141 & 1.026 & 5.291 & 0.215 & 0.816 & 0.945 & 0.979 \\
GeoNet~\cite{yin2018geonet} &  &  & K & 0.155 & 1.296 & 5.857 & 0.233 & 0.793 & 0.931 & 0.973 \\
EPC++~\cite{epc++} &  &  & K & 0.141 & 1.029 & 5.350 & 0.216 & 0.816 & 0.941 & 0.976 \\
SC-SfM~\cite{bian2019depth} &  &  & K & 0.119 & 0.857 & 4.950 & 0.197 & 0.863 & 0.957 & 0.981 \\
CC~\cite{competi_colab} &  &  & K & 0.140 & 1.070 & 5.326 & 0.217 & 0.826 & 0.941 & 0.975 \\
GLNet~\cite{glnet} &  &  & K & 0.135 & 1.070 & 5.230 & 0.210 & 0.841 & 0.948 & 0.980 \\
Monodepth2~\cite{monodepth2} &  &  & K & 0.115 & 0.882 & 4.701 & 0.190 & 0.879 & 0.961 & 0.982 \\
PackNet-SFM~\cite{packnet} &  &  & K & 0.111 & 0.785 & 4.601 & 0.189 & 0.878 & 0.960 & 0.982 \\
Lee \textit{et al.} \cite{lee2021attentive} &  & \checkmark & K & 0.112 & 0.777 & 4.772 & 0.191 & 0.872 & 0.959 & 0.982 \\
Gao \textit{et al.} \cite{gao} &  &  & K & 0.112 & 0.866 & 4.693 & 0.189 & 0.881 & 0.961 & 0.981 \\
FeatDepth~\cite{shu2020featdepth} &  &  & K & 0.109 & 0.923 & 4.819 & 0.193 & 0.886 & 0.963 & 0.981 \\
TC-Depth~\cite{ruhkamp2021attention} & \textbf{\checkmark} &  & K & {0.106} & 0.770 & 4.558 & 0.187 & 0.890 & 0.964 & 0.983 \\
RM-Depth~\cite{rmdepth} &  &  & K & 0.108 & \textbf{0.710} & 4.513 & 0.183 & 0.884 & 0.964 & 0.983 \\ 
Manydepth~\cite{manydepth} & \checkmark & & K & \underline{0.098} & 0.770 & \underline{4.459} & \underline{0.176} & \textbf{0.900} & \underline{0.965} & 0.983 \\
\textbf{Ours} & \checkmark &  & K & \textbf{0.094} & {\underline{0.734}} & \textbf{4.442} & \textbf{0.169} & \underline{0.893} & \textbf{0.967} & \textbf{0.983} \\ \hline
Struct2Depth~\cite{struct2depth} & \multicolumn{1}{l}{} & \checkmark & CS & 0.145 & 1.737 & 7.280 & 0.205 & 0.813 & 0.942 & 0.976 \\
Monodepth2~\cite{monodepth2} & \multicolumn{1}{l}{} &  & CS & 0.129 & 1.569 & 6.876 & 0.187 & 0.849 & 0.957 & 0.983 \\
Manydepth~\cite{manydepth} & \checkmark & \multicolumn{1}{l}{} & CS & 0.114 & 1.193 & 6.223 & 0.170 & 0.875 & 0.967 & 0.989 \\
InstaDM~\cite{lee2019instance} & \multicolumn{1}{l}{} & \checkmark & CS & 0.111 & 1.158 & 6.437 & 0.182 & 0.868 & 0.961 & 0.983 \\
Lee \textit{et al.} \cite{lee2021attentive} & \multicolumn{1}{l}{} & \checkmark & CS & 0.116 & 1.213 & 6.695 & 0.186 & 0.852 & 0.951 & 0.982 \\
RM-Depth~\cite{rmdepth} & \textbf{} & \multicolumn{1}{l}{} & CS & \underline{0.103} & \underline{1.000} & \underline{5.867} & \underline{0.157} & \underline{0.895} & \underline{0.974} & \underline{0.991} \\ 
\multicolumn{1}{l}{\textbf{Ours}} & \checkmark & \multicolumn{1}{l}{} & CS & \textbf{0.100} & \textbf{0.834} & \textbf{5.843} & \textbf{0.154} & \textbf{0.901} & \textbf{0.975} & \multicolumn{1}{c}{\textbf{0.992}} \\ \bottomrule
\end{tabular}
}
\end{table*}

\begin{table*}[t!]
\centering
\caption{\textbf{Depth Estimation Result on Dynamic Object Region.} We evaluate the depth prediction accuracy of dynamic objects (e.g.,Vehicles, Person, Bike) on KITTI and Cityscapes datasets. The instance masks are obtained using Maskformer~\cite{maskformer} pre-trained model.}
\label{tab:dynamic_result}
\scalebox{0.95}{
\begin{tabular}{lcccccccc}
\toprule
Methods & Dataset & AbsRel & SqRel & RMSE & RMSE$_{log}$ & $\delta < 1.25$ & $\delta < 1.25^2$ & $\delta < 1.25^3$ \\ \midrule
Monodepth2~\cite{monodepth2} & K & 0.172 & 1.892 & 5.732 & 0.273 & 0.819 & 0.917 & 0.945 \\
Packnet~\cite{packnet} & K & 0.169 & 1.873 & 5.797 & 0.276 & 0.809 & 0.913 & 0.942 \\
Insta-DM~\cite{lee2019instance} & K & 0.156 & 1.327 & 5.763 & 0.272 & 0.818 & 0.917 & 0.943 \\
Manydepth~\cite{manydepth} & K & 0.175 & 2.014 & 5.837 & 0.280 & 0.773 & 0.910 & 0.942 \\
\textbf{Ours} & K & \textbf{0.148} & \textbf{1.312} & \textbf{5.280} & \textbf{0.257} & \textbf{0.837} & \textbf{0.921} & \textbf{0.945} \\ \hline
Monodepth2~\cite{monodepth2} & CS & 0.159 & 1.948 & 6.492 & 0.217 & 0.820 & 0.948 & 0.981 \\
Packnet~\cite{packnet} & CS & 0.167 & 2.219 & 6.683 & 0.228 & 0.764 & 0.917 & 0.968 \\
Insta-DM~\cite{lee2019instance} & CS & 0.141 & 1.698 & 5.797 & 0.185 & 0.841 & 0.956 & 0.982 \\
Manydepth~\cite{manydepth} & CS & 0.163 & 2.138 & 6.537 & 0.220 & 0.770 & 0.928 & 0.971 \\
\textbf{Ours} & CS & \textbf{0.132} & \textbf{1.281} & \textbf{4.715} & \textbf{0.170} & \textbf{0.853} & \textbf{0.962} & \textbf{0.984} \\ \bottomrule
\end{tabular}
}
\end{table*}

\begin{figure*}[t]
\centering

\includegraphics[width=0.55\textwidth]{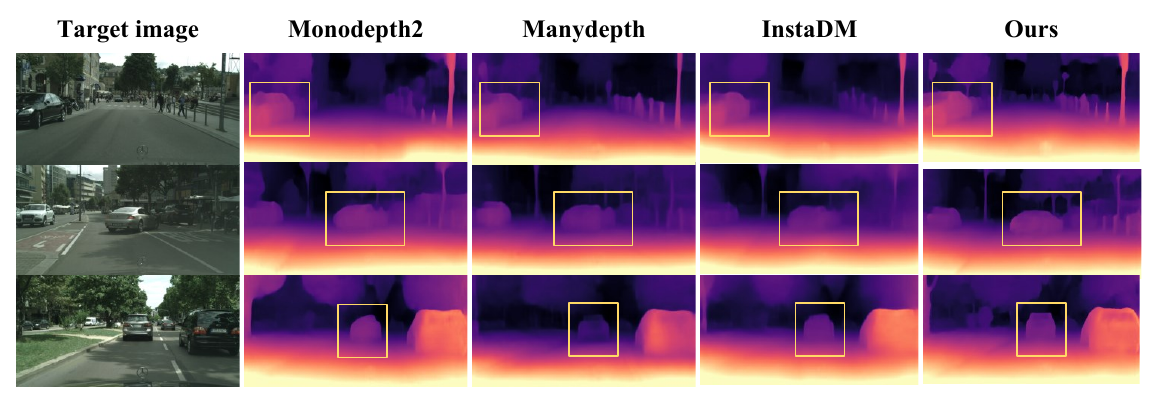}
\includegraphics[width=0.35\textwidth]{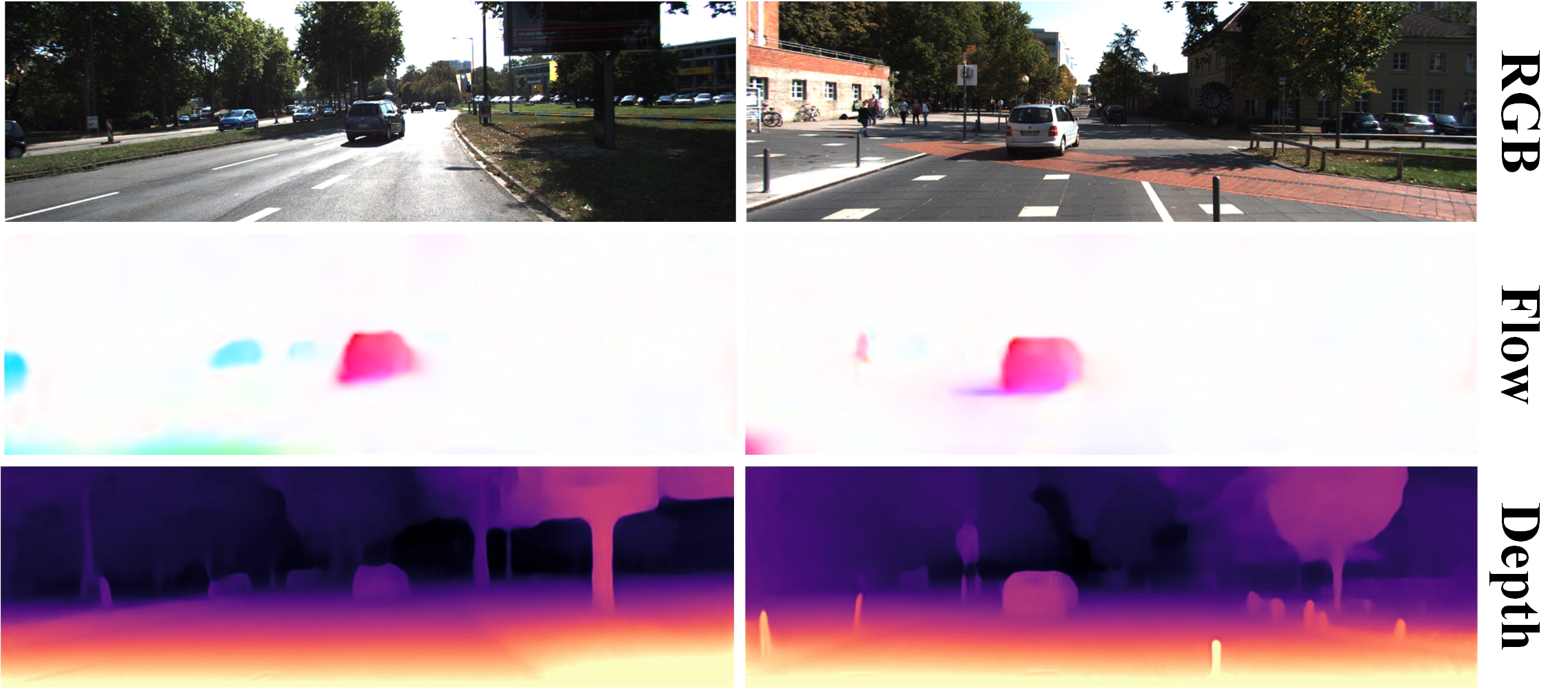}
\makebox[0.6\textwidth]{\small (a) Comparison of Depth Result }
\makebox[0.3\textwidth]{\small (b) Motion Field and Depth Result}

\caption{
\textbf{Qualitative depth estimation results} of our proposed Dyna-DepthFormer architecture, on the Cityscapes dataset. The results show that our method can achieve more accurate depth estimation results on dynamic objects and low-texture regions, while having sharper borders. In addition, our motion component can decouple object motion and camera ego motion efficiently.
}
\vspace{-4mm}
\label{fig:depth_result}
\end{figure*}

\begin{table}[t]
    \vspace{2mm}
    \centering
    \caption{Ablation study of attention modules on KITTI dataset.}
    \label{tab:cross_attention_ablation}
    \begin{tabular}{@{}lccccc@{}}
    \toprule
    \textbf{Method} & AbsRel & SqRel & RMSE & $\delta < 1.25$ & Mem (GB) \\ \midrule
    \textbf{full} & \textbf{0.100} & \textbf{0.698} & \textbf{4.236} & \textbf{0.903} & 6.6 \\
    W/o cross-attn & 0.106 & 0.714 & 4.562 & 0.887 & 3.8 \\
    W/o self-attn & 0.102 & 0.711 & 4.383 & 0.895 & 6.3 \\
    baseline & 0.108 & 0.721 & 4.421 & 0.883 & 3.5 \\ 
    \midrule
    K=1 & 0.108 & 0.774 & 4.823 & 0.882 & 3.7 \\
    K=2 & 0.106 & 0.746 & 4.548 & 0.887 & 4.6 \\
    K=4 & 0.100 & 0.714 & 4.320 & 0.895 & 7.8 \\
    \textbf{K=8} & \textbf{0.097} & \textbf{0.694} & \textbf{4.216} & \textbf{0.913} & 12.4 \\
    \bottomrule
    \end{tabular}
\end{table}



\begin{table}[t]
    \centering
    \caption{
    Ablation study of the motion field estimation network on the dynamic area in Cityscapes Dataset.
    \emph{Motion} indicates motion field based warping. \emph{Refine} indicates iterative refine step. \emph{Mask} indicates the auto-mask mechanism.
    }
    \label{tab:motion_ablation}
    \begin{tabular}{@{}cccccc@{}}
    \toprule
    Motion & Refine & Mask & AbsRel & RMSE & $\delta < 1.25$  \\ \midrule
     & & & 0.153 & 6.325 & 0.831 \\
    \checkmark & & & 0.142 & 5.835 & 0.837 \\
    \checkmark & \checkmark & & 0.138 & 5.331 & 0.847 \\
    \checkmark & & \checkmark & 0.136 & 5.218 & 0.850 \\
    \checkmark & \checkmark & \checkmark & \textbf{0.132} & \textbf{4.715} & \textbf{0.853} \\
    \bottomrule
    \end{tabular}
\end{table}

\begin{table}[t]
    \vspace{2mm}
    \centering
    \caption{Ablation study of the number of iterative refinement on dynamic area in Cityscapes dataset.}
    \label{tab:iter_num}
    \begin{tabular}{@{}ccccc@{}}
    \toprule
    \textbf{Iter} & AbsRel & SqRel & RMSE & $\delta < 1.25$  \\ \midrule
    1 & 0.136 & 1.576 & 5.342 & 0.849 \\
    2 & 0.133 & 1.410 & 4.922 & 0.852 \\
    4 & 0.132 & 1.293 & 4.723 & 0.853 \\
    6 & \textbf{0.132} & \textbf{1.281} & \textbf{4.715} & \textbf{0.853} \\ \bottomrule
    \end{tabular}
\end{table}

\subsection{Motion Estimate Network}
\label{sec:motion_estimation}
Previous research have tended to recover depth and camera motion simultaneously, but have ignored the non-rigid flow~\cite{yin2018geonet} caused by moving objects, resulting in incorrect view synthesis and a direct influence on the performance of self-supervised learning.
In this section, we propose a novel motion estimation network that aims to address moving objects problem in a general manner.
Specifically, We recover the camera and object motion with two separate heads, a motion field decode head and pose decode head, while they share the same motion feature encoder. 
Note that we do not utilize any semantic prior during training process.

The camera motion is also referred to as rigid flow. 
The shared encoder takes the image pair $\{\mathbf{I}_t, \mathbf{I}_s\}$ as input and outputs the relative pose $\{\mathbf{R}_{t \rightarrow s},\mathbf{t}_{t \rightarrow s}\}$ between them. 
Then, we can warp the source images $\mathbf{I}_s$ to target frame by differentiable view synthesis
\begin{equation}
    \mathbf{I}_{s \rightarrow t}=\mathbf{I}_{s}\left\langle\operatorname{proj}\left(\mathbf{D}_{t}, \{\mathbf{R}_{t \rightarrow s},\mathbf{t}_{t \rightarrow s}\}, \mathbf{K}\right)\right\rangle,
\end{equation}
where  $\left\langle \right\rangle$ is the bilinear interpolation sampling operator, $\operatorname{proj}()$ are the resulting 2D coordinates of the projected depths in $\mathbf{I}_s$, and $\mathbf{D}_t$ is the depth predicted by depth network.

As we synthesize the target view with the predicted camera motion, we suppose that the misaligned component is mainly caused by moving object.
Based on the above assumption, we then feed $\{\mathbf{I}_t, \mathbf{I}_{s \rightarrow t}\}$ into the shared encoder to generate object motion field $\textbf{T}_{res} \in \mathbb{R}^{3 \times H \times W}$. 
It is necessary to clarify that the motion field $\mathbf{T}_{res}$ is not exactly the same as scene flow \cite{teed2021raft3d}, as we consider the rotational motion of the moving object to be nearly zero.

\textbf{Iterative Refinement.} 
Our motion network are trained to decrease the local errors caused by object motion while learning the motion field.
However, due to the limitation of photometric consistency, motion variation arises during training.
To address this problem, we propose iterative refinement step to improve consistency and completeness of the predicted motion field. 
Specifically, we iteratively leverage $\textbf{T}_{res}$ and $\mathbf{I}_{t \rightarrow s}$ to reconstruct target image frame,
\begin{equation}
\begin{gathered}
\mathbf{I}_{s \rightarrow t}=\mathbf{I}_s\left\langle\operatorname{proj}\left(\mathbf{D}_t,\left\{\mathbf{R}_{t \rightarrow s}, \mathbf{t}_{t \rightarrow s}+\mathbf{T}_{\text {res }}\right\}, \mathbf{K}\right)\right\rangle \\
\Delta \mathbf{T}=\mathbf{M}_{\psi}\left(\mathbf{I}_t, \mathbf{I}_{t \rightarrow s}\right),~
\mathbf{T}_{\text {res }}=\mathbf{T}_{\text {res }}+\Delta \mathbf{T},
\end{gathered}
\end{equation}
where $\mathbf{T}_{res}$ is initialized to zero. By calculating the above process in several cycles, we can obtain the complete motion.


\subsection{Self-Supervised Objective}
In this section, we present the details of our self-supervised objective, which can be fomulated as:
\begin{equation}
   {L}=\lambda_p {L}_p + \lambda_{st} {L}_{st} + \lambda_{sd} {L}_{sd} + \lambda_{sp} {L}_{sp}.
\end{equation}
where ${\lambda_p,\lambda_{st},\lambda_{sd},\lambda_{sp}}$ are human-set parameters to balance the impact of each loss function on the model training,


\textbf{Photometric consistency loss.}
The photometric loss measures the difference between the target image $\mathbf{I}_t$ and the synthesized image $\mathbf{I}_{s \rightarrow t}$ for self-supervised training. The photometric error is defined as a combination of a L1 and SSIM loss:
\begin{equation}
    L_p = \alpha \frac{1-\operatorname{SSIM}\left(\mathbf{I}_{t}, \mathbf{I}_{s \rightarrow t}\right)}{2}+(1-\alpha)\left\|\mathbf{I}_{t}-I_{s \rightarrow t}\right\|_{1},
\end{equation}
where $\alpha$ is set to $0.85$. 
To address the occlusion pixel problem, we compute the per-pixel minimum reprojection loss across all source views rather than averaging them, and we also integrate the auto-masking mechanism to filter dynamic pixel.

\textbf{Smoothness regularization.}
Following \cite{monodepth2}, we also use an edge-aware depth smoothness regularization and depth smoothness regularization to improve the performance in low texture regions:

\begin{equation}
    {L}_{st}=\sum_{p}\left(\nabla \mathbf{T}_{res}(p) \cdot e^{-\nabla \mathbf{D}_t(p)}\right)^{2}
\end{equation}

\begin{equation}
    L_{sd}=\sum_{p}\left(\nabla \mathbf{D}_{t}(p) \cdot e^{-\nabla \mathbf{I}_{t}(p)}\right)^{2}
\end{equation}

\textbf{Motion Field Sparsity loss.}
In order to encourage the sparsity of motion field prediction, we also design a motion field regularization loss:

\begin{equation}
    {L}_{sp}= \sum_{p} L_{g1}(\mathbf{T}_{res}),
\end{equation}
where $L_{g1}()$ is chosen to be the sparsity function as it is mentioned in \cite{dynamic_scenes}.

\label{sec:loss_function}

\section{EXPERIMENTS}

\subsection{Datasets}
\textbf{KITTI}~\cite{geiger2013vision} is an autonomous driving dataset collected in urban and highway areas. We adopted the pre-processing method of \cite{zhou2017unsupervised} to remove the static frames in Eigen split~\cite{eigen2014depth}. 
Finally, The training set includes 39,810 image triples, while the validation set has 4,424 images.
The depth map ground truth is generated by projecting the LiDAR pointcloud into the camera coordinate.
Note that, we only calculate accuracy and error metrics on these sparse depth points.
The results shown are tested on 697 test images following the steps in \cite{garg2016unsupervised}.

\textbf{Cityscapes}~\cite{cityscapes} is also a popular urban driving dataset.
The training set contains 69,730 image triplets, and the validation set includes 1,525 images.
We used the cropping and evaluation protocol mentioned in~\cite{laina2016deeper}.

\subsection{Implementation Details}
\textbf{Network Details.} We designed Depth Estimation Network with an encoder and decoder structure based on ResNet50 \cite{he2016resnet}, which is pre-trained with ImageNet \cite{deng2009imagenet}.
The decoder has the same structure as HR-Depth ~\cite{lyu2021hrdepth}, and its output is a multi-scale inverse depth map.
The transformer module is realized follow by \cite{zhu2020deformable}. The total attention head M is set to 8, and the total sample points K is set to 4 for efficiency. 

\textbf{Training Details.} Our models are implemented using PyTorch~\cite{paszke2017automatic} and trained across four NVIDIA GTX 2080~Ti GPUs.
We use the Adam optimizer~\cite{kingma2014adam} , with $\beta_1 = 0.9$ and $\beta_2=0.999$, and a batch size of 4 per GPU. 
The loss weights $\{\lambda_p,\lambda_{st},\lambda_{sd},\lambda_{sp}\}$ are set to $\{1, 0.01, 0.001, 0.01\}$.
We train our networks for 25 epochs, with the learning rate of 2e-4 for the first 15 epochs and reduce the learning rate to 2e-5 for the rest epochs.
To ensure the stability of the training process, we freeze the pose estimation network after 15 epochs. 


\subsection{Benchmark Performance}
In order to validate our Dyna-DepthFormer architecture, we performed an in-depth comparison of its performance against other published methods.

Table~\ref{tab:depth_result} shows the depth prediction results on the Cityscapes \cite{cityscapes} and KITTI \cite{gtkitti} testing sets. 
We compare with three types of methods: monocular depth estimation methods which predict the depth without using multi-frame information in the same mini-batch, and cost-volume based method including Manydepth~\cite{manydepth} and TC-Depth~\cite{ruhkamp2021attention}, and dynamic object modeling methods including InstaDM~\cite{lee2019instance}, struct2depth~\cite{struct2depth} and RM-Depth~\cite{rmdepth}, etc.
As shown in Table~\ref{tab:depth_result}, Dyna-DepthFormer outperforms the compared methods on both KITTI and Cityscapes datasets.
Cost-volume based methods can achieve promising result, while they suffer from feature matching noise on dynamic area, which has a large impact on depth estimation accuracy.
Differently, we fuse the multi-view features through a more robust cross-attention based approach and therefore achieve more accurate depth estimation results.
Compared to the KITTI dataset, the Cityscapes dataset involves more moving objects, therefore the improvement of our method on the cityscapes dataset is more significant. 

Table~\ref{tab:dynamic_result} shows the depth prediction on dynamic objects region (e.g.,Vehicles, Person, Bike).
We obtain the dynamic instance masks using the pre-trained state-of-the-art instance segmentation method~\cite{maskformer}. 
To cope with dynamic objects during training process, we did not use two-stage training scheme~\cite{lee2021attentive} or apply instance-level semantic prior, but our approach achieved better results by decoupling object and ego motion explicitly. 
Qualitative results are shown in Fig.~\ref{fig:depth_result}.

\subsection{Ablation Analysis}
\textbf{Effectness of Attention Module.} 
As shown in Table~\ref{tab:cross_attention_ablation}, the complete model outperforms the baseline model by a large margin. The proposed attention components are effective in improving the depth accuracy. By removing either the cross attention or self attention module, the depth accuracy is decreased. This is to be expected, because without them, models can only predict the depth from a single 
view feature map, rather than estimating depth by matching features between adjacent frames.
We also modified the parameters of our architecture and evaluated its performance, obtained by modifying the number of sampling points K. 
By observing the trends in the Table~\ref{tab:cross_attention_ablation}, we can learn that increasing the complexity of the cross-attention network leads to better results. 
Also, it is further demonstrated that robust feature matching is beneficial for depth estimation task.

\textbf{Effectness of Motion Field Prediction.}
As shown in Table~\ref{tab:motion_ablation}, utilization of the motion field for warping-based view synthesis significantly improves depth estimate accuracy in dynamic regions. 
In addition, we optimize the prediction of motion field by iterative refinement. As shown in Table \ref{tab:iter_num}, the accuracy of dynamic object depth estimation generally improves as the number of repetitions grows, but ultimately approaches saturation.

\section{Conclusion}
This paper proposes a novel attention-based architecture for self-supervised monocular depth estimation in dynamic scenes. 
First, we designed a set of attention-based modules to perform feature aggregating across adjacent images efficiently, which has proved to be helpful in improving the overall performance of our network. 
Second, we designed an object motion field estimation network to model the dynamic object explicitly without leveraging any semantic prior. The effectiveness of our algorithm has been demonstrated on various autonomous driving datasets.
Future work will consider the use of the proposed method for vision-based SLAM system in dynamic environments.


\bibliographystyle{ieeetr}
\bibliography{ref}

\end{document}